\documentclass[conference]{IEEEtran}
\IEEEoverridecommandlockouts
\usepackage{cite}
\usepackage{amsmath,amssymb,amsfonts}
\usepackage{algorithmic}
\usepackage{graphicx}
\usepackage{textcomp}
\usepackage{xcolor}
\def\BibTeX{{\rm B\kern-.05em{\sc i\kern-.025em b}\kern-.08em
    T\kern-.1667em\lower.7ex\hbox{E}\kern-.125emX}}

\usepackage{booktabs}
\usepackage{multirow}
 
\usepackage{soul}

\usepackage{svg}
\usepackage{hyperref}

\begin{document}

\title{Parameter Efficient Hybrid Transformer (PEHT) for Network Traffic Prediction via Dynamic Urban Congestion Integration\\
\thanks{This work was supported in part by the National Science Foundation (NSF) under Grant No. CNS-2543209; by the U.S. Department of Transportation (USDOT) Tier-1 University Transportation Center (UTC), Transportation Cybersecurity Center for Advanced Research and Education (CYBER-CARE), under Grant No. 69A3552348332; and by the NVIDIA Academic Grant Program, which provided GPU resources used in this research.}
}

\author{\IEEEauthorblockN{1\textsuperscript{st} Abdolazim Rezaei}
\IEEEauthorblockA{\textit{Department of Computer Science} \\
\textit{Texas A\&M University}\\
Corpus Christi, USA \\
arezaei@islander.tamucc.edu}
\and
\IEEEauthorblockN{2\textsuperscript{nd} Mehdi Sookhak}
\IEEEauthorblockA{\textit{Department of Computer Science} \\
\textit{Texas A\&M University}\\
Corpus Christi, USA \\
m.sookhak@ieee.org}
\and
\IEEEauthorblockN{3\textsuperscript{rd} Mahboobeh Haghparast}
\IEEEauthorblockA{\textit{Department of Computer Science} \\
\textit{Texas A\&M University}\\
Corpus Christi, USA \\
mahboobeh.haghparast@tamucc.edu}
}

\maketitle

\begin{abstract}

Accurate network traffic prediction is a critical element for efficient resource allocation in dynamic urban cellular networks. However, prediction remains challenging because network demand is influenced by complex mobility patterns, congestion dynamics, and heterogeneous user behavior. This paper introduces the Parameter-Efficient Hybrid Transformer (PEHT), a network traffic prediction framework that integrates urban mobility and congestion information into a Transformer-based architecture. PEHT separates primary network communication features from secondary urban mobility features and incorporates Low-Rank Adaptation (LoRA) into the Transformer encoder to reduce the number of trainable parameters while maintaining high predictive accuracy. A multimodal fusion strategy then injects external mobility and congestion features into the decoder to improve traffic forecasting. Experiments on the Telecom Italia Milan dataset and multiple synthetic congestion scenarios show that PEHT outperforms state-of-the-art baselines in terms of RMSE, MAE, and $R^2$. The implementation is available in the \href{https://github.com/azim015/PEHT}{GitHub} repository.
\end{abstract}


\IEEEpeerreviewmaketitle

\section{Introduction}

% Network traffic forecasting has been a component for efficient resource management in modern cellular, Wi-fi and etc infrastructures. 
Network traffic forecasting is a key component of efficient resource management in modern cellular and wireless infrastructures.
Accurate predictions are important for optimizing network performance which can manage resource allocation and it ensures connectivity in increasingly dynamic environments. 
However, the complexity of urban traffic considering diverse user behaviors and rapidly changing environmental factors and demands resulted in challenges for more accurate prediction models.

Traditional network traffic prediction models struggle to capture the external factors and other data sources that influence traffic patterns. 
While recent advancements in Graph Neural Networks (GNNs) \cite{wu2022graph, liu2020towards, li2025kolmogorov} and Transformer-based models \cite{liang2025interaction, bharilya2025self, huang2025tailored} have shown promise in learning spatio-temporal representations, they frequently show limitations. 
GNNs face scalability limitations despite their ability to model graph-structured data, and Transformer models become computationally and parameter-intensive when applied to high-dimensional data, while being excellent at capturing long-range dependencies.
%GNNs, despite their ability to model graph-structured data, can suffer from scalability issues due to high-order message passing and the over-smoothing problem, leading to indistinguishable region representations in deeper architectures. Similarly, standard Transformer models, while excellent at capturing long-range dependencies, may become computationally intensive and parameter-heavy when applied to the high-dimensional, real-world traffic data generated by a multitude of mobile devices across a city. These limitations hinder their practical deployment and their ability to generalize effectively to uncertain and dynamic urban conditions.

To address these challenges, we introduce a Parameter-Efficient Hybrid Transformer (PEHT) architecture. Our PEHT model is specifically designed to significantly improve network traffic prediction by integrating real-time urban mobility and congestion data, which are often overlooked or inadequately incorporated by existing methods. The core innovation lies in its ability to efficiently handle high-dimensional input spaces without compromising predictive performance.

Our approach divides input features into two main groups of features including primary network communication features and secondary urban mobility features such as traffic flow and congestion. The primary features are processed by a customized Transformer Encoder which is enhanced with Low Rank Adaptation (LoRA). Unlike standard LoRA used for fine-tuning foundation models, we apply this low-rank decomposition ($W = AB^T$) during the initial training phase. This acts as a powerful regularizer for the noisy Milan traffic data, preventing overfitting while reducing the trainable parameter count by over 90\%. 
%LoRA enables efficient handling of the high-dimensional spatio-temporal and communication data from numerous mobile devices by adapting only a small and low-rank subset of the Transformer's parameters. This significantly reduces the number of trainable parameters and makes the model computationally feasible for large-scale urban deployments. 
The secondary urban mobility features are then robustly fused with the \textit{Encoder}'s output which provides the \textit{Decoder} module with rich contextual cues regarding urban dynamics affecting traffic periodicity.

Experimental results demonstrate that the proposed PEHT model outperforms previous state-of-the-art approaches. It utilizes the Transformer model to incorporate external information that directly impacts urban traffic dynamics. Our key contributions are summarized as follows:
\begin{itemize}
    \item We propose PEHT for network traffic prediction expecting to integrate diverse urban mobility and congestion data for network traffic prediction.

    \item We introduce the application of LoRA within the Transformer Encoder to efficiently handle high-dimensional spatio-temporal network traffic data.
    %, significantly reducing model parameters without performance degradation.

    \item We develop a robust multimodal fusion strategy that injects real-time urban mobility features into the Transformer's decoding process which provides explicit insights. % for more accurate future traffic predictions.

%    \item We conduct comprehensive experiments on real-world datasets demonstrating that PEHT achieves superior prediction accuracy and computational efficiency compared to existing methods.
    
\end{itemize}

%The remainder of this paper is organized as follows: Section 2 reviews related work in network traffic prediction. Section 3 details the proposed PEHT methodology, including device-to-base station association, multimodal data aggregation, the customized Transformer model, and external information fusion. Section 4 presents and discusses our experimental results which includes implementation details, performance comparisons, and ablation studies. Finally, Section 5 concludes the paper and outlines directions for future work.

\section{Related Work}

Network traffic prediction has been widely studied in recent years. Based on the types of features and tasks, researchers have mainly used two types of methods: GNN-based models and Transformer-based models. Below, the most recent studies from each group are discussed in more detail. 

\subsection{Graph-based Models}
Proposed by Chen et al. \cite{chen2023novel}, this model captures spatio-temporal characteristics to provide accurate predictions. It consists of a spatial module using GCNN for spatial characteristics based on spatial distance and correlation, and a temporal module using three LSTM networks for temporal characteristics based on temporal proximity and periodicity. 
%The results from both modules are then fused. The model demonstrates significant improvements in RMSE, MAE, and R-squared values compared to advanced models on real cellular traffic datasets.

Yang et al. \cite{yang2025multi} propose a framework which addresses node heterogeneity, complex traffic propagation, and periodic dependencies in IoT networks. It integrates a Multi-Representation Graph Convolutional Network (MRGCN) module to model dynamic connectivity patterns through adaptive adjacency matrices and node heterogeneity via differentiated graph convolution paths. 
%It also includes Historical Embedding and Temporal Embedding modules to capture and fuse periodic dependencies across different fine-grained temporal cycles..

Mehrabian et al. in \cite{mehrabian2023dynamic} for wireless cellular traffic prediction, considering spatial, temporal, and spectral domain information. It includes a Spectral Dynamic Graph Construction (SDGC) method to model spatial dependencies between cells using Discrete Fourier Transform (DFT) and GRU in the frequency domain, along with a self-attention mechanism. 
%A Dynamic Bernstein Polynomial Filtering (DBPF) scheme is developed to capture spatial correlations and infer cell-specific parameters using K-order Bernstein polynomial approximation. DBGRN integrates DBPF with a GRU network to predict spatio-temporal traffic demands. Experiments show DBGRN outperforms baseline schemes in RMSE and MAE.

Zhang et al. \cite{zhang2024cellular} propose HGCRN which combines a static graph convolutional recurrent neural network and meta-graph learning to capture complex spatio-temporal dependencies. It also constructs graph adjacency matrices that go beyond geographical proximity by integrating distance-based construction with meta-graph learning. 
%HGCRN uses an encoder-decoder architecture with GCRN units (LSTM and GCN) to extract spatio-temporal features. The model demonstrates significant improvements in MAE, MAPE, and RMSE on real-world datasets.

In another study, Guo et al. \cite{guo2023capturing} propose AGCN to address the "adjacent but not dependent" and "dependent but not adjacent" problems. It uses a combination of GAT and GCN to capture embedded spatial correlations, and a dilated convolution module with a gate mechanism to efficiently learn long-term temporal relations. 
%This model focuses on network topology information in traffic prediction for each node. AGCN shows excellent performance in prediction accuracy and inference time compared to mainstream methods.

\subsection{Transformer-based Models}
Liu et al. \cite{liu2021st} introduced ST-Tran, a novel model that simultaneously explores spatial and temporal sequence information in cellular traffic. It features a temporal Transformer block for recent and periodic temporal features, and a spatial Transformer block to capture spatial characteristics from related grids. 
%ST-Tran specifically addresses the challenge of insufficient information from scalar traffic data by using complementary input sequences and incorporating data augmentation and grid selection to improve accuracy. Their work demonstrated the usability of Transformer structure in cellular traffic prediction for the first time.

Gong et al. \cite{gong2025sttf} proposed STTF for multi-task mobile network prediction, specifically for mobile traffic and connected user amount. STTF incorporates a temporal cross-attention encoder to capture the complex interaction between mobile traffic and connected users, and a hierarchical spatial encoder to identify relevant information from various semantic relationships.
%(e.g., proximity, function similarity, pattern similarity, flow similarity). 
%A subgraph sampling method is utilized to reduce computational power requirements, making it applicable to large-scale networks. STTF significantly outperforms state-of-the-art models in both mobile traffic and connected user prediction. The cross-attention mechanism in STTF is highlighted as being more effective for feature exchange between mobile traffic and user data compared to parameter sharing in other multi-task methods.

In another study, Guan et al. \cite{guan2024st} introduced ST-DCAN, a Transformer-based model designed to address model complexity and the extraction of global information in traffic prediction. ST-DCAN integrates a unified attention mechanism on input patches and employs dual compression attention to reduce computational complexity.
%from $O(N^2)$ to $O(N)$
%for both temporal and spatial dimensions. This allows for efficient extraction of global temporal (intra-node) and spatial (inter-node) dependencies. ST-DCAN shows state-of-the-art performance on real-world traffic forecasting datasets with optimal computational efficiency.

Shuvro et al. \cite{shuvro2023transformer} utilized a modified Transformer architecture called 2D-Transformers, for time-series vehicular data to predict traffic flow in Software Defined Network (SDN)-based Vehicular Ad-hoc Networks (VANETs). Their model captures inter-feature correlations along with inter-sample correlations using a 2D Multi-head Attention Mechanism. 
%This approach aims to provide traffic guidance by passing learned model parameters through the SDN-enabled network. The 2D-Transformers model showed a significant decrease in error compared to traditional Transformers and LSTM-based models, especially in capturing sudden changes in traffic flow.

Habib et al. \cite{habib2024transformer} introduced a Transformer-based method, specifically Autoformer , for predicting wireless network traffic in concise temporal intervals for Open Radio Access Networks (O-RAN). The system dynamically launches reinforcement learning-based traffic steering xApps or cell sleeping rApps based on predicted traffic volume to optimize performance metrics like throughput and energy efficiency. 
%Autoformer's decomposition architecture and auto-correlation mechanism are leveraged to handle intricate temporal patterns and inherent periodicity in traffic data for accurate forecasting. This prediction-based on-demand activation approach offers significant improvements in energy efficiency and throughput compared to "always on" applications.

In addition, Yan et al. \cite{zhang2023st2t} proposed ST2T for cellular traffic prediction within Digital Twin (DT) systems. ST2T combines convolutional and self-attention mechanisms in a hybrid encoder and employs a two-stage decoder to handle the high dynamic range of spatio-temporal data. 
%It also utilizes tubelet patching for intuitive capture of adjacent space-time relationships, departing from traditional 2D patches. ST2T has demonstrated substantial advancements in cellular traffic prediction within DT systems, outperforming various baselines by effectively leveraging both spatial and temporal dependencies.

Despite promising results by recent studies, critical limitations remain across both model families. GNN-based approaches encounter scalability challenges in dense urban environments and Transformer-based models become parameter intensive when they receive high-dimensional data especially when it is spatio-temporal. Additionally, neither category adequately incorporates external urban mobility and congestion signals that directly influence network traffic patterns, and those that attempt integration often risk violating temporal causality during inference. These gaps motivate the design of a better framework which addresses parameter inefficiency and bridges the external data integration gap.

\section{Methodology}
%\input{3_Proposed_Methodology}

%\section{Methodology}
\subsection{Spatio-Temporal Grid Clustering and Virtual Cell Construction}
The initial step discusses the spatial mismatch between raw user mobility and the available aggregated data. Since the Telecom Italia dataset provides aggregated traffic per grid square $g \in G$ rather than user-level traces, we apply a Grid-to-Virtual-Cell mapping strategy. Adjacent grid squares are clustered into Virtual Base Stations ($VBS$) using a customized greedy clustering algorithm. This resembeles the service area coverage of physical towers and mitigates the data sparsity often found in high resolution grid squares which creates more robust signals for the prediction model. For a grid square $g$ with traffic time series $X_g$, the greedy algorithm assigns it to a Virtual Base Station cluster $v^* \in V$ that maximizes the utility function $U(g,v)$. Unlike traditional clustering, this function integrates spatial dependence and load similarity to improve statistically meaningful grouping such that

\begin{equation}
v^* = \arg \max_{v \in V} U(g,v)
\end{equation}
The utility function is formulated to balance spatial proximity with temporal correlation:
\begin{equation}
U(g,v) = \alpha \cdot Corr(X_{g},X_{v}) - \beta \cdot Dist(g,v) - \gamma \cdot |Load_{g} - \overline{Load}_{v}|
\end{equation}

where $\mathrm{Corr}(X_g,X_v)$ represents the Pearson correlation coefficient between the traffic pattern of grid square $g$ and the centroid traffic pattern of cluster $v$. This encourages grouping grid squares with similar temporal behavior, such as commercial or residential traffic patterns.
The term $\mathrm{Dist}(g,v)$ represents the Euclidean distance between the grid center and the cluster center, while $\left|\mathrm{Load}_g-\overline{\mathrm{Load}}_v\right|$ penalizes high load variance within a cluster. The coefficients $\alpha$, $\beta$, and $\gamma$ are weighting factors that control the relative importance of temporal similarity, spatial proximity, and load balance, respectively. Following this association, the traffic $T_{v}$ for a Virtual Base Station $v$ is computed by aggregating the traffic of its constituent grids: $T_{v} = \sum_{g \in G_{v}} T_{g}$, where $G_v$ denotes the set of grid squares assigned to virtual base station $v$. This aggregation forms the foundational multivariate time series used by the prediction model, smoothing high-frequency noise from individual grid measurements while preserving regional urban mobility dynamics.

\subsection{Multimodal Data Aggregation and Feature Engineering}

a feature vector is computed for each grid square $g$ to create a comprehensive input for the prediction model. This vector combines the spatial distance of the grid centroid to the virtual base station center with the aggregated traffic volume which includes SMS, Call, Internet recorded in that square. These grid-level features are then aggregated to form the multivariate time series input for each Virtual Base Station ($VBS$).

Also, the model uses various data modalities to capture complex network dynamics. These include Historical Traffic Data ($X_{\text{hist}}$), Temporal Features ($F_{\text{temp}}$), Static Network Features ($F_{\text{static}}$), Dynamic Real-Valued Features ($F_{\text{dynamic}}$).
% \begin{itemize}
%     \item \textbf{Historical Traffic Data ($X_{\text{hist}}$):} The primary time series input, representing past aggregated traffic volumes for each base station over a \texttt{context\_length} period.
%     \item \textbf{Temporal Features ($F_{\text{temp}}$):} Encoding cyclical patterns (e.g., hour of the day, day of the week, holiday indicators). These serve as positional encodings for time series data.
%     \item \textbf{Static Network Features ($F_{\text{static}}$):} Unchanging characteristics of base stations (e.g., unique base station IDs, geographical coordinates, cell type, road network topology in the coverage area). These provide static spatial context.
%     \item \textbf{Dynamic Real-Valued Features ($F_{\text{dynamic}}$):} Time-varying network state variables (e.g., current number of connected devices, interference levels).
% \end{itemize}

All these diverse inputs are transformed into a unified numerical representation, typically high-dimensional vectors, suitable for the Transformer's attention mechanism. Continuous numerical values are converted via a "vector-to-vector" embedding layer, $\text{Embed}_{\text{cont}}(\cdot)$. Categorical features are mapped to dense vectors using learnable embedding lookup tables, $\text{Embed}_{\text{cat}}(\cdot)$.

\textbf{Positional Encoding} ($PE(t)$) is added to the input embeddings to inject information about the relative or absolute position of each data point within the time series, as Transformers process sequences in parallel without inherent order understanding. For an input embedding $E_t$ at time step $t$, the positional encoding $PE(t)$ is added $E_{\text{final},t} = \text{Embed}(x_t) + PE(t) $
A common sinusoidal positional encoding is defined as
\begin{equation}
PE(t,2i)=\sin\left(\frac{t}{10000^{2i/d_{\mathrm{model}}}}\right)
\end{equation}
and 

\begin{equation}
PE(t,2i+1)=\cos\left(\frac{t}{10000^{2i/d_{\mathrm{model}}}}\right)
\end{equation}

where $t$ is the position (time step), $i$ is the dimension index, and $d_{\text{model}}$ is the embedding dimensionality.

\begin{figure}[h!]
    \centering
    \includegraphics[width=0.95\linewidth]{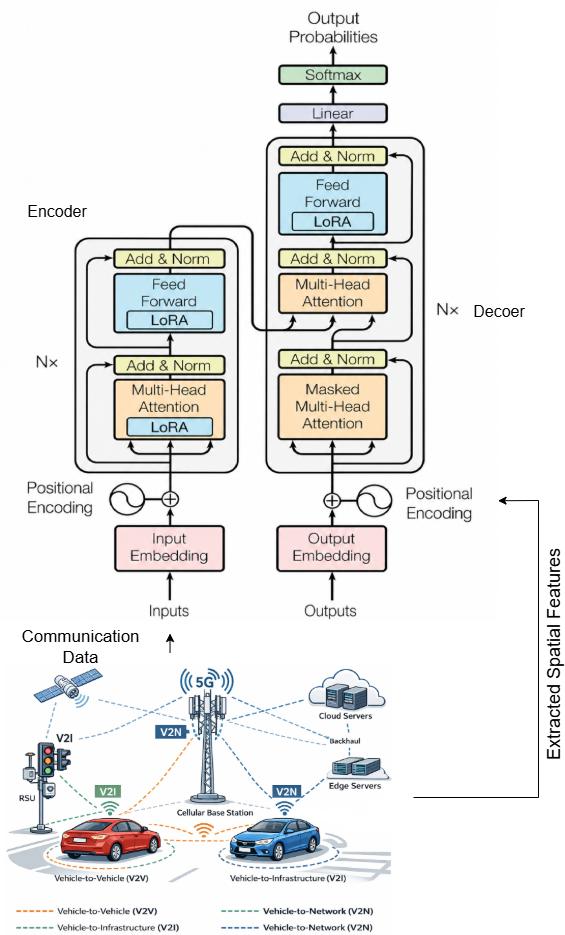}
    \caption{PEHT: The proposed framework initially applies LoRA on \textit{Encoder} and merges with vehicle mobility features to be fed into \textit{Decoder}.}
    \label{fig:main}
\end{figure}

The fusion of these modalities can happen at different stages. For instance, \textbf{feature fusion} includes extracting features from each modality and then concatenating these vectors before feeding them into the Transformer's main blocks. If $E_{\text{hist},t}$, $E_{\text{temp},t}$, $E_{\text{static}}$, and $E_{\text{dynamic},t}$ a features at time $t$ (or static for $E_{\text{static}}$), the fused input vector $I_t$ for the Transformer at time $t$ can be represented as:
\begin{equation}
I_t = \text{Concat}(E_{\text{hist},t}, E_{\text{temp},t}, E_{\text{static}}, E_{\text{dynamic},t})
\end{equation}
This rich feature set, especially the spatiotemporal features, is critical for the Transformer to learn the complex interplay between different network entities and their traffic patterns.

\subsection{Customized Transformer Model: The Prediction Core}

The core of the prediction model is a customized Transformer architecture which is adapted for continuous time series data. The proposed Transformer would be capable on the multi-head attention mechanism which allows for parallel processing of input sequences and it will be effective to capture long-range dependencies, and finally overcomes limitations of traditional recurrent networks.

The model employs an \textbf{Encoder-Decoder structure}:
\begin{itemize}
    \item \textbf{Encoder:} Processes a context\_length of historical traffic values and their associated multi-modal features (e.g., past\_values, past\_time\_features, static\_categorical\_features, static\_real\_features). Its role is to learn a rich, contextualized representation of the input historical sequence, capturing both temporal and spatial patterns. If the input sequence to the encoder is $X_{\text{enc}} \in \mathbb{R}^{L_{\text{enc}} \times d_{\text{model}}}$, the encoder produces an output $H_{\text{enc}} \in \mathbb{R}^{L_{\text{enc}} \times d_{\text{model}}}$.
    
    \item \textbf{Decoder:} Predicts a prediction\_length of future traffic values (future\_values) based on the encoder's output, potentially incorporating known future covariates like future\_time\_features. The decoder typically operates autoregressively, generating predictions step-by-step, where previously predicted values are fed back as input for subsequent steps.
\end{itemize}

\textbf{LoRA:}
Given the high spatial resolution of the Milan grid network and the resulting high-dimensional feature space, \textbf{LoRA} is devised as a parameter-efficient method. LoRA modifies a pre-trained model by adjusting only a small, low-rank subset of its parameters, significantly reducing the number of trainable parameters. Instead of updating the entire weight matrix $W$ of a layer, LoRA approximates the weight update $\Delta W$ by the product of two low-rank matrices, $A$ and $B$ such that 

\begin{equation}
\Delta W \approx AB 
\end{equation}
where \[A \in \mathbb{R}^{d_{\text{out}} \times r}\]

\noindent and
$B \in \mathbb{R}^{r \times d_{\text{in}}}$, with $r$ being the low rank ($r \ll \min(d_{\text{in}}, d_{\text{out}})$). 

% This means that instead of training $d_{\text{in}} \times d_{\text{out}}$ parameters for $\Delta W$, LoRA trains $d_{\text{out}} \times r + r \times d_{\text{in}}$ parameters, a significant reduction for small $r$. The matrices $A$ and $B$ can be initialized, for example, by performing Singular Value Decomposition (SVD) on the original weight matrix $W$ and taking the top $r$ singular values and vectors: $W \approx U_r S_r V_r^T$, then $A = U_r \sqrt{S_r}$ and $B = \sqrt{S_r} V_r^T$. 

% The adapted output $y'$ for an input $x$ becomes:
% $$y' = Wx + \alpha (AB)x$$
% Here, $\alpha$ is a scaling hyperparameter that controls the influence of the low-rank adaptation, balancing between retaining pre-trained knowledge and adapting to the new task. This approach allows for efficient fine-tuning of the large Transformer model, making it computationally feasible for real-world deployment with massive data.

\subsection{Fusion of External Traffic Information}

After the encoder processes the aggregated grid traffic and network features, its output $H_{\text{enc}}$ (a rich, contextualized representation) is fused with \textbf{extra traffic-related information} ($F_{\text{external}}$). 
%This external data includes macroscopic insights like the number of vehicles in each base station's coverage area. This fusion step is critical for a holistic prediction, allowing the model to leverage both fine-grained device-level patterns and broader contextual information that influences network traffic.
%
The fusion can be implemented as a feature fusion strategy where the encoder's output vector is concatenated with the external traffic features. 
%If $H_{\text{enc},t}$ is the encoder output for time step $t$, and $F_{\text{external},t}$ is the external traffic feature vector for time step $t$, the fused data $X_{\text{fused},t}$ fed into the decoder is:
%$X_{\text{fused},t} = \text{Concat}(H_{\text{enc},t}, F_{\text{external},t})$
%This combined, fused data is then fed into the decoder.

To ensure strict temporal causality, the fused external features $F_{external}$ correspond strictly to time steps $t$ (historical) or are derived from autoregressive forecasts ($\hat{t+k}$),ensuring no ground-truth future information is leaked to the decoder during the inference phase.

\subsection{Prediction}

The \textbf{Decoder} receives the fused data $X_{\text{fused}}$ (encoder output combined with external traffic information) and performs the final prediction. It leverages the learned contextual representations and the additional traffic insights to generate the predicted network traffic for each base station over the specified future prediction\_length. The decoder's masked self-attention ensures that predictions for a given time step only depend on known past data, maintaining causality. The final output layer of the decoder typically consists of a linear transformation to project the decoder's hidden states to the desired output dimension:
$\hat{Y} = W_{\text{out}} X_{\text{dec\_out}} + b_{\text{out}}$
 where $\hat{Y}$ is the predicted traffic sequence, $X_{\text{dec\_out}}$ is the output from the final decoder layer, and $W_{\text{out}}, b_{\text{out}}$ are the weight and bias of the output layer.

This multi-stage model, from intelligent device assignment and comprehensive feature engineering to the efficient, attention-driven Transformer core with LoRA, is designed to provide accurate and actionable network traffic forecasts, enabling proactive optimization and resource management in dynamic cellular environments. The proposed framework is shown in Figure~\ref{fig:main}.

\section{Results and Discussions}
\subsection{Experiments}

\begin{table*}[htbp]
\centering
\caption{Network Traffic Prediction Results on Milan Dataset Considering Spatiotemporal Features Using PEHT.}
\label{tab:results_milan}
\renewcommand{\arraystretch}{1.2}
\begin{tabular}{ll|ccc|ccc|ccc}
\toprule
\multirow{2}{*} & \multirow{2}{*}{\textbf{Methods}} 
& \multicolumn{3}{c|}{\textbf{SMS}} 
& \multicolumn{3}{c|}{\textbf{Call}} 
& \multicolumn{3}{c}{\textbf{Internet}} \\
\cline{3-11}
& & RMSE$\downarrow$ & MAE$\downarrow$ & R2$\uparrow$ 
  & RMSE$\downarrow$ & MAE$\downarrow$ & R2$\uparrow$
  & RMSE$\downarrow$ & MAE$\downarrow$ & R2$\uparrow$ \\
\midrule

\multirow{7}{*}{} 
& HGCRN \cite{zhang2024cellular}     & 75.2 & 30.28 & -  & 75.2 & 30.28 & -                         & 75.2 & 30.28  & - \\
& ST-DenseNet \cite{zhang2018citywide} & 43.9073 & 19.6701  & 0.8437 & 60.3758 & 31.3021 & 0.8191 & 196.3721 & 125.0611 & 0.9290 \\
& STCNet  \cite{zhang2019deep}    & 34.3346 & 17.9901 & 0.9102      & 54.1664 & 30.3221 & 0.8590 &     167.3321 & 93.8873  & 0.9500 \\
& StTran  \cite{liu2021st}    & 39.8221 & 16.4982 & 0.9300      & 56.5428 & 33.4029 & 0.8198 &         169.7015 & 91.3930  & 0.9489 \\
& MVSTGN  \cite{yao2021mvstgn}    & 30.9443 & 14.6816 & 0.9379      & 49.0515 & 24.9796 & 0.8856 &     165.0445 & 88.6983  & 0.9550 \\
& ST2T \cite{zhang2023st2t}     & 21.5861 & 14.4490 & 0.9352      & 40.0087 & 23.9413 & 0.8594 &       138.6490 & 79.6580  & 0.9562 \\
% & Customized CNN \cite{zhu2024adaptive}   & 23.0323 & 14.6816 & -     & 23.0323 & 14.6558 & -       & 23.0323 & 14.6558  & - \\
% & CNN LSTM (2D) \cite{fu2022traffic}  & 9.2705 & \textbf{3.2975} & 0.9344 & 15.8464 & 5.1503 & 0.6659        & 135.5715 & 61.7303 & 0.9072 \\
\midrule

& \textbf{PEHT (ours)} 
& \textbf{18.42} & 12.15 & \textbf{0.9481} 
& \textbf{34.85} & \textbf{20.60} & \textbf{0.8842} 

& \textbf{122.10} & \textbf{68.45} & \textbf{0.9685} \\
\bottomrule
\multicolumn{11}{p{0.85\textwidth}}{\footnotesize \textit{$^{\mathrm{*}}$ Note: The performance of the HGCRN baseline is reported based on the official implementation using default hyperparameters. Its relatively high RMSE in our experiments is attributed to the specific temporal train/test split utilized, which challenges the model's adaptability to non-stationary distribution shifts compared to the Transformer-based baselines.}} \\
\end{tabular}
\end{table*}

To conduct a robust and reliable experimental analysis, we perform the experiment from "Telecom Italia Big Data Challenge (Milan)" \cite{barlacchi2015multi} to compare our model’s performance with previous studies. 
Additionally, We conducted experiments using the Telecom Italia Milan dataset~\cite{barlacchi2015multi} and generated five synthetic datasets for ablation studies under varying congestion and mobility conditions.

\begin{figure}[htbp]
    \centering
    \includegraphics[width=3.1in]{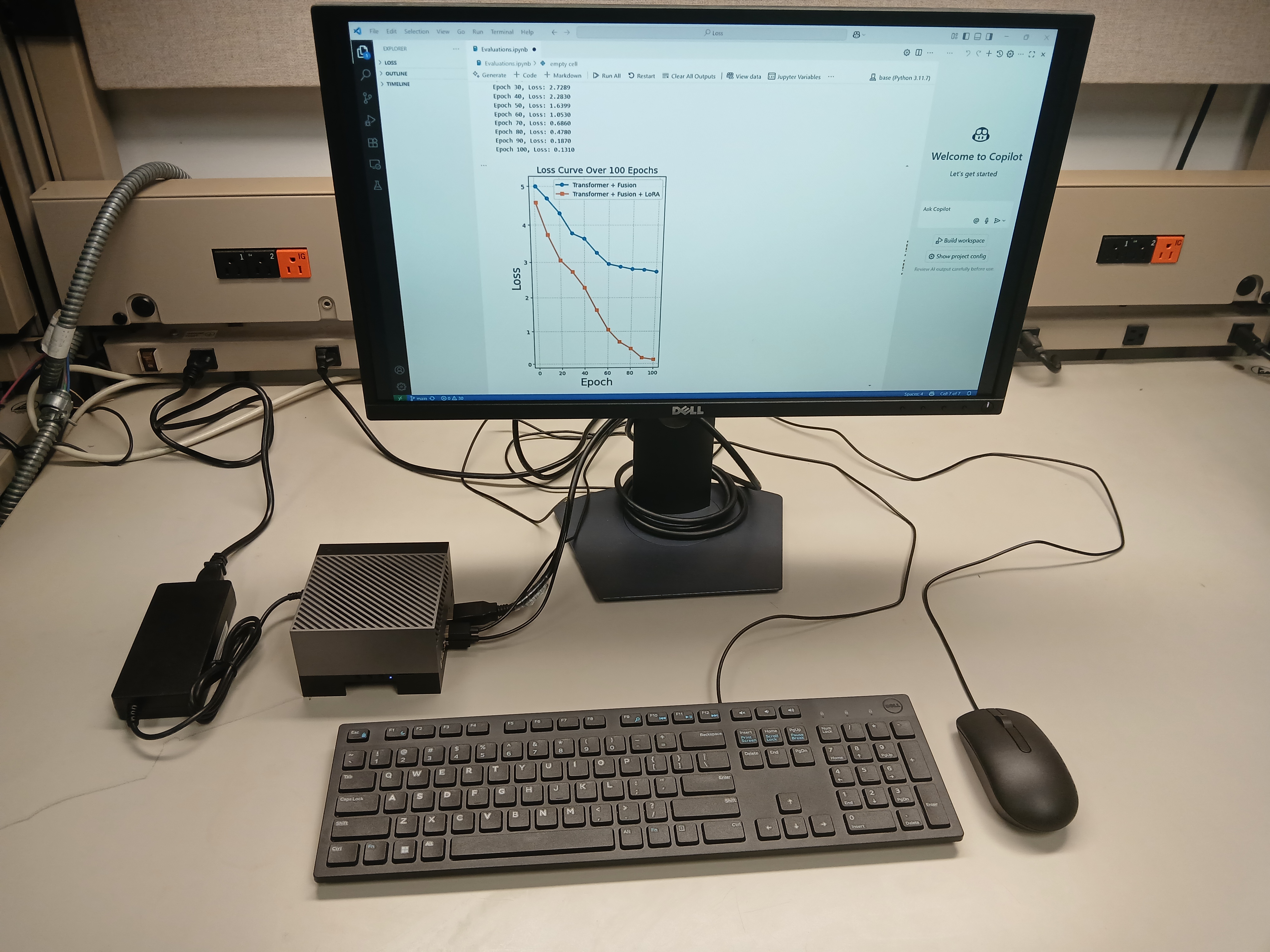}
    \caption{NVIDIA Jetson AGX Orin Development Kit and supporting equipment used for experimental deployment.}
    \label{fig:jetson}
\end{figure}

The PEHT model was trained using the AdamW optimizer with a learning rate of $5\times10^{-4}$ and a weight decay of $1\times10^{-4}$. The input sequence length was set to 48 time steps, while multiple prediction horizons were evaluated for future traffic forecasting. The Transformer architecture employed a hidden dimension of 128, four attention heads, three encoder layers, and two decoder layers. A dropout rate of 0.1 was applied to mitigate overfitting. For the LoRA module, the rank was set to $r=8$ with a scaling factor of $\alpha=16$, providing a balance between prediction accuracy and parameter efficiency. The model was trained for 100 epochs using early stopping based on the validation loss.

As shown in Table \ref{tab:results_milan}, PEHT considerably outperforms the state-of-the-art models such as ST2T as the baseline model. Specifically, PEHT achieves a 14.6\% reduction in RMSE for SMS traffic (18.42 vs. 21.58) and an 11.9\% reduction for Internet traffic. This performance gain confirms that the low-rank constraint effectively regularizes the model against high-frequency noise, while the decoder-fusion strategy successfully integrates urban mobility dynamics without violating temporal causality.

% C_Congested
\begin{table}[htbp]
\centering
\caption{Results on Five Synthetic Datasets with different conditions.}
\label{tab:synthetic}
\renewcommand{\arraystretch}{1.2}
\begin{tabular}{|ll|ccc|}
\toprule
\multirow{2}{*} & \multirow{2}{*}{\textbf{Datasets / Methods}} 
& \multicolumn{3}{c|}{\textbf{Metrics}} \\
\cline{3-5}
& & RMSE$\downarrow$ & MAE$\downarrow$ & R2$\uparrow$ \\
\midrule
\multirow{4}{*}{C\_Congested} 
& Encoder Fusion     &  13.1171  & 8.6847 & 0.8391  \\ 
& LoRA    & 10.5287 & 6.6246 & 0.8929  \\
& \textbf{Full Model} & \textbf{9.2437} & \textbf{5.4972} & \textbf{0.9051}  \\

\midrule
\multirow{2}{*}{E\_Congested}
& Encoder Fusion  & 12.6157 & 7.2549 & 0.8652  \\
%\midrule
& LoRA & 10.0107 &  4.5562 & \textbf{0.9286}  \\
& \textbf{Full Model} & \textbf{8.4827} & \textbf{3.9728} & 0.9203  \\

\midrule
\multirow{2}{*}{w/o\_Ped}
& Encoder Fusion  & 10.5922 & 5.7327 & 0.9316  \\
%\midrule
& LoRA & 9.3632 & \textbf{3.8829} & 0.9568  \\
& \textbf{Full Model} & \textbf{9.1938} & 4.0162 & \textbf{0.961}  \\

\midrule
\multirow{2}{*}{w/o\_V}
& Encoder Fusion  & 10.1506 & 4.3194 & 0.8781  \\
%\midrule
& LoRA & 9.7554 & 4.5481 & 0.9017  \\
& \textbf{Full Model} & \textbf{9.1196} & \textbf{4.1005} & \textbf{0.9182}  \\

\midrule
\multirow{2}{*}{Regular}
& Encoder Fusion  & 11.332 & 4.9549 & 0.8803  \\
%\midrule
& LoRA & 11.006 & 4.1510 & 0.9551  \\
& \textbf{Full Model} & \textbf{8.8510} & \textbf{3.3306} & \textbf{0.9709}  \\

\bottomrule
\end{tabular}
\end{table}

\subsection{Ablation Study}
In addition to the Telecom Italia Milan \cite{barlacchi2015multi} dataset, five synthetic datasets were generated using the CARLA simulator to evaluate the performance of the proposed model under diverse mobility and network conditions across three stages: train/validation/test with 70/20/10 percent of the entire data. 
These scenarios include base-station congestion (C\_Congested), cell-edge congestion (E\_Congested), vehicular-only mobility (w/o\_Ped), pedestrian-only mobility(w/o\_V), and regular mobility patterns (Regular). Each synthetic dataset isolates a specific traffic characteristic to assess model behavior under controlled conditions. 
%We applied our model on five different synthetic datasets representing different conditions. 
The model is also available in three versions, with the Encoder Fusion version enhanced by data fused with the Encoder. LoRA version represents the model with the LoRA module, and the Encoder's output is passed directly to the Decoder. The Full model performs with all features presented in the proposed architecture. As shown in Table~\ref{tab:synthetic}, the Full Model achieves the best or near-best performance across most synthetic scenarios, demonstrating the complementary benefit of combining encoder fusion with LoRA. Although LoRA alone achieves the highest $R^2$ in the cell-edge congestion scenario, the Full Model provides the most consistent overall performance across RMSE, MAE, and $R^2$.

\begin{figure}[htbp]
    \centering
    \includegraphics[width=3.6in]{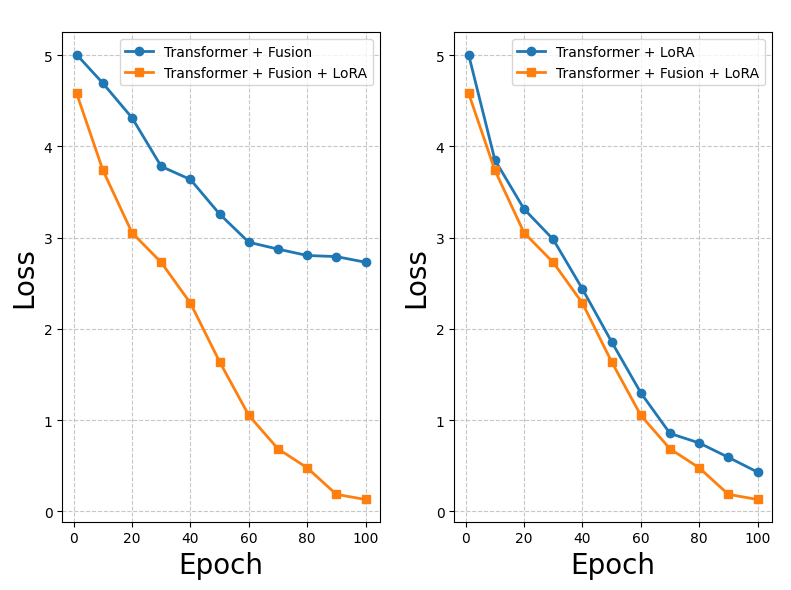}
    \caption{Ablation comparison of the Full Model against variants excluding encoder fusion or LoRA.}
    \label{fig:loss}
\end{figure}

The loss values are considered to evaluate the Full model against the Encoder Fusion and LoRA modules. Figure \ref{fig:loss} shows the impact of LoRA in the model, which is more significant than the Encoder Fusion. 

The LoRA method significantly reduces the number of parameters in the PEHT model by approximating a large weight matrix update with the product of two much smaller, low-rank matrices. This approach makes the model computationally feasible for large-scale urban deployments.

To quantify the parameter reduction, we consider an input dimension of 4,372 and LoRA ranks $r\in\{4,8,16\}$. Therefore, for the full weight matrix with an input and output dimension of $4372$, the original number of parameters to be trained would be approximately $4372 \times 4372 \approx 19.1 million$.
However, by applying LoRA with a low-rank parameter $r$, the number of trainable parameters is drastically decreased. The LoRA approach trains $(d_{\text{out}} \times r) + (r \times d_{\text{in}})$
parameters instead of the full $d_{\text{in}} \times d_{\text{out}}$ parameters.

To quantify the parameter reduction, consider input and output dimensions of 4,372. A full weight matrix contains approximately $4{,}372 \times 4{,}372 \approx 19.1$ million parameters. In contrast, LoRA trains only $(d_{\mathrm{out}}\times r)+(r\times d_{\mathrm{in}})$ parameters. For ranks $r=4$, $r=8$, and $r=16$, this corresponds to 34,976, 69,952, and 139,904 trainable parameters, respectively, representing a substantial reduction compared with the full matrix.

% Here is how the parameter count changes with different values of $r$:

% \begin{itemize}
%     \item For a low rank of $r = 4$, the number of trainable parameters would be approximately:
%     $$4372 \times 4 + 4 \times 4372 = 34976$$,
%     which is a massive reduction from 19.1 million.

%     \item For a rank of $r = 8$, the parameters increase to:
%     $$4372 \times 8 + 8 \times 4372 = 69952,$$
%     still a tiny fraction of the original count.

%     \item Even with a rank of $r = 16$, the trainable parameters are only $4372 \times 16 + 16 \times 4372 = 139904,$   representing a significant reduction.
% \end{itemize}

\section{Conclusion}
This paper presents the PEHT architecture, a novel approach designed to enhance network traffic prediction by integrating real-time urban mobility and congestion data. The innovation in the PEHT model is a customized Transformer Encoder which is enhanced with LoRA to efficiently handle high-dimensional input from numerous mobile devices without compromising performance. Another novelty is the fusion of vehicle mobility features with the Encoder's output before it is passed to the Decoder module. This provides the Decoder with explicit contextual insights into future traffic conditions. Experimental results indicate that the PEHT model outperforms previous approaches by using the Transformer to model complex network behavior while efficiently incorporating crucial external information.


\ifCLASSOPTIONcompsoc
  \section*{Acknowledgments}
\else
% \fi

% \hl{The authors would like to thank...}

\bibliographystyle{IEEEtran}

%\bibliography{\jobname}
\bibliography{References}



% \end{thebibliography}

\end{document}